\documentclass[preprint,12pt,numbers,sort&compress]{elsarticle}

\usepackage{setspace}
\usepackage{geometry}
\usepackage[table,xcdraw,svgnames]{xcolor}
\usepackage{soul}
\usepackage{colortbl}
\usepackage{tcolorbox}

\usepackage{amsmath}
\usepackage{amssymb}
\usepackage{mathtools}
\DeclareMathOperator*{\argmin}{argmin}

\usepackage{graphicx}
\usepackage{caption}
\usepackage{subcaption}
\usepackage[figuresright]{rotating}
\usepackage{here}
\usepackage{float}
\graphicspath{{Figures/}}

\usepackage{tabularx}
\usepackage{multirow}

\let\oldtabular=\tabular
\def\tabular{\small\oldtabular}

\usepackage{framed}
\usepackage{lipsum}
\usepackage{upgreek}
\usepackage[bottom]{footmisc}
\usepackage{enumitem}
\usepackage{lineno}

\usepackage{algorithm}
\usepackage{algpseudocode}

\usepackage[
    colorlinks=true,
    linkcolor=blue,
    urlcolor=blue,
    citecolor=blue,
    anchorcolor=blue
]{hyperref}

\journal{arXiv}

\begin{document}

\begin{frontmatter}



\title{\textbf{Capabilities of Auto-encoders and Principal Component Analysis of the reduction of microstructural images; Application on the acceleration
of Phase-Field simulations
}   }



\author[a]{Seifallah Fetni} 
\author[a]{Thinh Quy Duc Pham}
\author[c]{Truong Vinh Hoang}
\author[b]{Hoang Son Tran}
\author[a]{Laurent Duchêne}
\author[b]{Xuan-Van Tran}
\author[a]{Anne Marie Habraken}\corref{mycorrespondingauthor1}

\cortext[mycorrespondingauthor1]{Corresponding author e-mail: s.fetni@uliege.be}

\address[a]{University of Liège, UEE Research Unit, MSM division, allée de la Découverte, 9 B52/3, B 4000 Liège, Belgium}
\address[b]{Institute of Strategy Development, Thu Dau Mot University, 75100 Binh Duong Province, Viet Nam}
\address[c]{Chair of Mathematics for Uncertainty Quantification, RWTH-Aachen University, 52056 Aachen, Germany}

\begin{abstract}
In this work, a data-driven framework based on Phase-Field simulations data is proposed to highlight the capabilities of neural networks to ensure accurate low dimensionality reduction of simulated microstructural images and to provide time-series analysis.
The dataset was indeed constructed from high-fidelity Phase-Field simulations. Analyses demonstrated that the association of auto-encoder neural networks and principal component analyses leads to ensure efficient and significant dimensionality reduction : 1/196 of reduction ratio with more than 80 \% of accuracy. These findings give insight to apply analyses on data from the latent dimension. Application of Long Short Term Memory (LSTM) neural networks showed the possibility of making next frame predictions ; that makes possible the acceleration of Phase-Field simulation without the need of high computing resources. We discussed the application of such a framework on various areas of research. Different methods are proposed from the conducted analyses, in order to ensure dimensionality reduction (auto-encoders, principal component analysis, Artificial Neural Networks) and time-series analysis (LSTM, Gated Recurrent Unit (GRU)).
\end{abstract}
\end{frontmatter}
\tableofcontents
\tolerance=1
\emergencystretch=\maxdimen
\hyphenpenalty=10000
\hbadness=10000
\section{Introduction}
\par Phase-Field method (PFM) is considered as a promising tool for highly non linear problems related to different fields of mechanics such as heat transfer, fluid flow, moving boundaries and crystalline anisotropy. In particular, it is well adapted for studying  solidification kinetics based on the driving forces, phase transformations and moving boundaries and prediction of the microstructure evolution and the solidified materials. Interesting application to study solidification phenomena \cite{GU2021110812,LINDROOS2022103139}, crack propagation \cite{BOISSE20076151} and the coarsening of precipitates \cite{LI2022111259,WANG2022104246} can be highlighted.
Therefore, PFM was developed to figure from the most promising tools used in computational materials science to deal with microstructural changes in materials under high thermal gradients, and evolved to be integrated in Computational Materials Engineering (ICME) that has as main purpose the improvement of materials design \cite{AAGESEN201710}. 
\par Through the last decades, Artificial Intelligence (AI) has been noticeably evolved and applied in various areas : business, engineering, medicine, entertainment \cite{KUMBHAR20215467,RABBANI2022}..
Machine Learning (ML) is defined as the set of techniques that lead to realize AI. When dealing with materials science, ML has attracted attention and been more and more applied to resolve various problems. Deep Learning (DL) is a type of ML based on deep neural networks in order to progressively extract high level of data features. 
Artificial Neural Networks (ANN) frameworks have offered a first approach as surrogate models of various kinds of constitutive laws for linear and non-linear materials .e.g. elasto-plastics and thermal field prediction in Additive Manufacturing processes \cite{Pham2021,Fetni2021esaform, PHAM2022103297}, to predict texture evolution and stress–strain response using based-crystal plasticity frameworks \cite{IBRAGIMOVA2021103059}, development of new steels \cite{BHATTACHARYYA20131}. 
ANN frameworks have been indeed used in supervised learning when the targets are known and already figure in the training dataset. However, in the case of unlabelled data, where the outputs are unknown, unsupervised learning algorithms should be applied. One can enumerate some works where classifier such as support vector machine (SVM) were used to classify solid phases in low carbon steels \cite{GOLA2019186} or ferrito-pearlitic and martensitic-austenitic steels using random forest operators \cite{GUPTA2020107224}. One should note that Convolutional Neural Networks (CNN) have proven efficiency in classification problems \cite{FIRAT2022100694}.
\newline With regards to supervised learning, some limitations were reported with the use of ANNs, in particular when dealing with highly non-linear materials with irreversible behaviours. Another drawback is when the time-dependent relationship between the actual and the precedent step is not taken into account. They can not indeed deal with time-series analysis, which consists one of the main advanced topics in materials science such as Phase-Field, fatigue crack propagation and crystal plasticity.
This aspect constitutes a net limitation of ANNs to be applied as replacement of constitutive laws for visco-plasticity. 
This issue is mainly due to the sequential aspect of the information involved in the plastic deformations.
Therefore, alternative approaches must be explored \cite{ABUEIDDA2021102852} in order to  accurately model the complex behaviour of history-dependent materials. To handle the sequential information, Recurrent Neural Networks (RNNs) were recently introduced in the AM field. They provided promising results when predicting unknown results such as stresses and plastic energy \cite{WU2020113234}. This architecture allows correlations between temporal and non-temporal input parameters and enables the update of the information between the previous inputs and the future prediction. 
Therefore, alternative approaches must be explored \cite{ABUEIDDA2021102852} in order to  accurately model the complex behaviour of history-dependent materials. 
But a major drawback should be highlighted; the problem of vanishing or exploding gradients with RNNs. Indeed, when the chain is long, the output gradient becomes too small and weights can not anymore be updated \cite{REHMER20201243}. Therefore, training RNN models becomes very difficult in this case. In order to overcome this drawback, Long Short Term Memory (LSTM) neural networks have been proposed as replacements of RNNs. New gates are introduced in the chain in order to overcome the vanishing or exploding gradients. The idea of such architecture is to divide the signal between what is important in the short term through the hidden state (analogous to the output of a single RNN cell), and what is important in long term, through the cell state (newly introduced). In the most recent ML applications on AM and other physical domains, LSTM as well as other sequential neural networks (Temporal Convolutional Networks (TCN) \cite{MEKA2021119759} and Gated recurrent units (GRU) \cite{SHI2020104122}) provided encouraging performances to accurately predict the history-dependent responses. Combined CNN-LSTM has been proposed to deal with the time-series problems \cite{HUANG2022106685}
\par However, there are still few works which have been focused on the time-series problems related to microstructures in the literature\cite{SMontes2021}. This is essentially due to the high amount of computing resources needed to achieve the training of the models on the time-series datasets. Data should be first compressed before neural network processing. In this respect, principal Component Analysis (PCA) is a popular tool to compress microstructure-based data. That can be ensured by capturing the key components of the microstructure then putting them in a linear space and gathering the redundant information representing the microstructure \cite{XU201739,LATYPOV2018671}. In their work, de Oca Zapiain et al. \cite{SMontes2021} proposed a framework applied on dataset of Phase-Field simulated microstructures. After applying two-points statistics and PCA for low representation of the microstructure, they (\cite{SMontes2021}) applied LSTM on the reconstructed image to accelerate Phase-Field simulations. This idea gives new insights to the capabilities of Neural Network to deal with the computing time of heavy simulations such as Phase-Field and crystal plasticity. The abundance of microstructural images in the literature can enhance researches in this direction. 
Meanwhile, some issues should be addressed with regard to the applied method for dimensionality reduction to represent microstructures in latent spaces. 
Indeed, autocorrelation-based methods such as PCA representation does not fit the highly non-linear dependencies of microstructural images, even if this limitation could be addressed by Phase-Field simulations (when fed by the predicted next frame issued from LSTM, PFM has the ability to compute the predicted microstructure and to evolve it). Such method of reduction indeed doesn't allow the obtaining of near net-shapes images as inferred from results of \cite{SMontes2021}. 
In their recent work, Hu et al. \cite{HU2022115128} compared PCA to Time-aligned isometric mapping method (Isomap) to deal with dimensionality reduction. They found that Isomap can not scale well with large datasets. Another method was also tested : Uniform manifold approximation and projection (UMAP). They mentioned that it should be preceded by PCA to handle Phase-Field data. 
In order to adapt the framework for other applications (next frame prediction, failure time estimation, in-situ calibration of process parameters ..), near-net shape images should be obtained after reduction and reconstruction. To address this issue, non-linear dimensionality-reduction techniques are highly recommended, in particular artificial neural network-based methods \cite{HU2022115128}. For that, auto-encoders could be proposed as potential candidates for achieving a reduced dimensional representation of microstructural images prepared for further investigations \cite{KO2022116094,CHEVROT2022102652}. This tool indeed ensures data compression and decompression with neural networks. That makes them good candidates for data denoising and dimensionality reduction \cite{ALAHMADI2022102658}. Another important advantage of the use of auto-encoders is that they are automatically trained on samples from the built dataset. The combination of auto-encoder with another method for dimensionality reduction (e.g. PCA) could be investigated.
\par
\begin{figure*}
\centering
\subcaptionbox{ \label{}}
{\includegraphics[width=14.5cm,height=7cm]{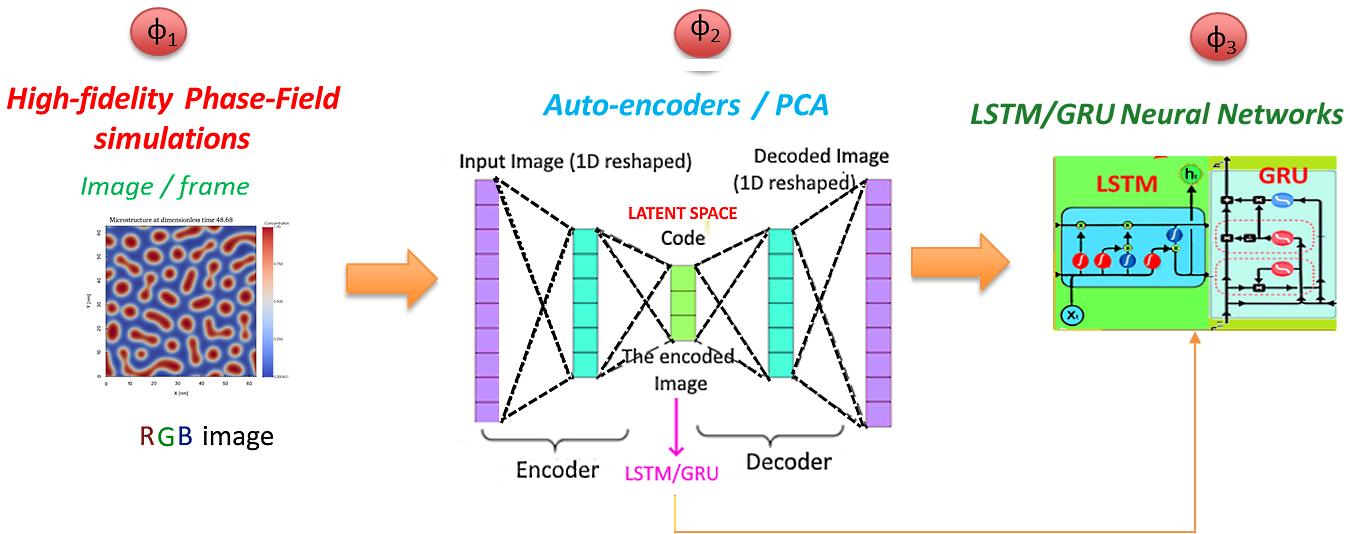}}
\subcaptionbox{ \label{}}
{\includegraphics[width=10cm,height=5cm]{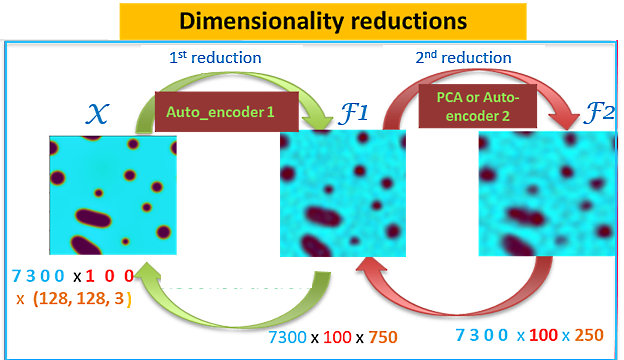}}
\subcaptionbox{ \label{}}
{\includegraphics[width=14cm,height=5cm]{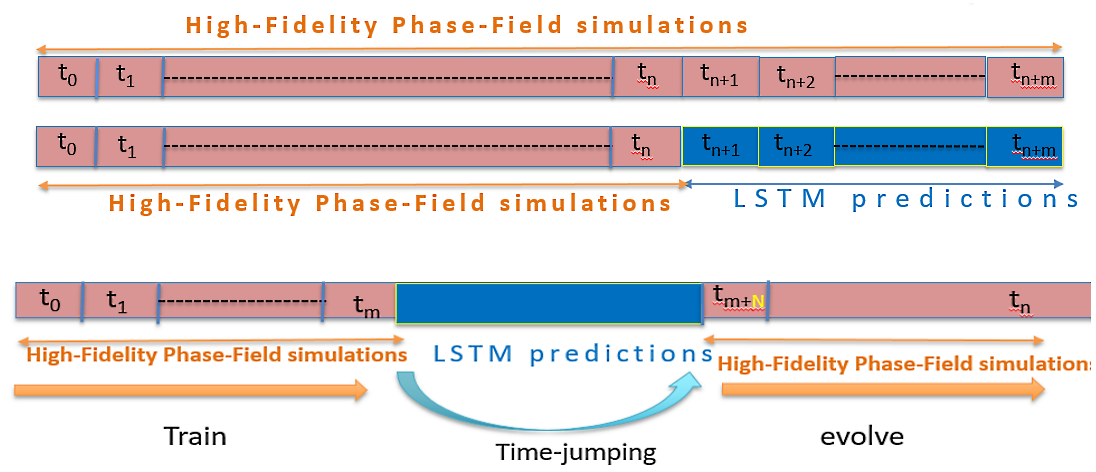}}
\caption{ Machine-learning based framework (surrogate model) for the dimensionality reduction of Phase-Field microstructure images and accelerating of phase-field based microstructure simulations. (a)  (left) Phase-Field simulations to build the training and validation datasets, (middle) Low-dimensional representation based on Auto-encoders and PCAs, (right) Time-series analysis: LSTM and GRU for comparison purposes. (b) Illustration of the double layer reduction : the layer 1 consists of an auto-encoder neural network while the layer 2 is an  auto-encoder or a PCA. (c) Possible scenarios to apply the proposed frameworks to accelerate Phase-Field simulations}
\label{fig:flowchart}
\end{figure*}

 In this work, a new framework is proposed based on auto-encoders neural networks. It consists of 3 phases compared to the 4 ones proposed in \cite{SMontes2021}, as follows. First, the dataset is constructed by running a big number of Phase-Field simulations using parallel computing techniques. Then, auto-encoders are trained on several examples from the built dataset. Two configurations are compared: a double layer of auto-encoders on one hand and a combination Auto-encoder-PCA on the other hand.
The reduced representation (called hereafter codes) are used as time observations in LSTM neural networks. Finally, once trained, the third step results in predicted frames/images instantly reconstructed to give near-net shape images. 
The framework is shown in Fig. \ref{fig:flowchart}. Here $\phi$1, $\phi$2 and $\phi$3 denote the three phases of the framework. 

\section{Methods}
\subsection{Phase Field-based dataset}
\begin{figure}
\centering
\subcaptionbox{ \label{}}
{\includegraphics[width=9cm,height=6cm]{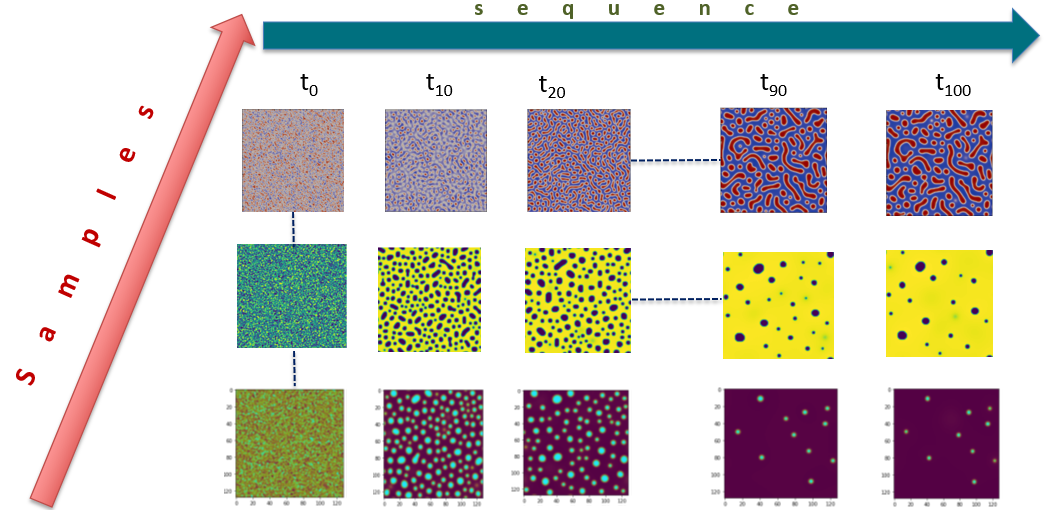}}
\caption{Examples of the microstructure evolutions of binary alloys under spinodal decomposition for different values of \textit{X}, \textit{M} and $\kappa$.}
\label{fig:dataset}
\end{figure}

The microstructure evolution for the spinodal decomposition of a given binary alloy can be studied by tracking the concentration \textit{X} of the alloying elements. It is governed by the Cahn-Hilliard equation as follows: 
\begin{equation}
\frac{\partial X}{\partial t}=\nabla M.\frac{\partial G}{\partial X} 
\label{eq:eq1}
\end{equation}
Here \textit{G} represents the total free energy of the binary alloy, \textit{X} the concentration (molar fraction) of one element (Si in AlSi10Mg for example) and \textit{M} is the mobility (in $J^{-1} m^{-1} s^{-1}$). 
\textit{G} can be expressed when neglecting the terms related to the elastic strain as follows:
\begin{equation}
G(X)=\int_{}^{} ( g(X)+\frac{1}{2} \kappa.(\nabla X)^2) dV
\label{eq:eq2}
\end{equation}
$\kappa$ (in $J. m^{-1}$) is the gradient energy coefficient and \textit{g} is the chemical bulk energy ; it can be expressed as a 4$^{th}$ polynomial in function of \textit{X}:
\begin{equation}
 g(X)=AX^2 (1-X^2)
\label{eq:eq3}
\end{equation}
where \textit{A} is a constant. Indeed, this coefficient controls the energy barrier height between the equilibrium phases. It is here chosen equal to 1 for simplification purposes. 
\newline In order to generate a large dataset, we choose to vary the concentration, the mobility and the gradient energy coefficient in the ranges 0.25:0.75, 0.8:2.2 and 0.25:0.75 respectively. Indeed, as the purpose of the work is mainly numerical (investigating the capabilities of promising neural networks to deal with time-series problems in materials science), the target dataset should be as diverse as possible. 
Therefore, we generate more than 10000 sets of parameters (\textit{X}, \textit{M}, $\kappa$), to feed the auto-encoder and LSTM neural networks models. For computing purposes, only 7000 sets are randomly chosen to train the models (hereafter described). 
For the generation of the dataset, spinodal decomposition simulations were performed using a 2D square grid with a uniform
mesh of 128 × 128 x 3 grid points (RGB representation) dimensionless spatial. The spatial discretization is set as  $\Delta$ x = $\Delta$ y = $\Delta$ z= 1 , while the temporal discretization $\Delta$t = 0.01 (dimensionless). Each simulation was obtained by running 20000 time steps under periodic boundary conditions and the semi-implicit spectral method was applied. More details about spinodal decomposition simulations are given in \cite{Fetni2021COMPLAS}, where a link to the open-source gitlab repository is provided containing the scripts for reproducing the work.
\newline Generation of the dataset and the encoding work (hereafter explained) was conducted in the Clusters NIC5, Hercules2 and Dragon2 of  the CÉCI (Consortium des Équipements de Calcul Intensif) ; (cf. section acknowledgement). These resources allow advanced computing, e.g. NIC5 consists of 4672 cores spread across 73 compute nodes, with two 32 cores AMD Epyc Rome 7542 CPUs at 2.9 GHz. The default partition holds 70 nodes with 256GB of RAM. Once the dataset is encoded, the training of the LSTM model could be ensured with an ordinary machine (e.g. i7-10510U. 8 MB cache, up to 4.90 GHz). 
The virtual experiments generated the dataset which is saved as a Numpy 3D array as : (\textit{samples $\times$ sequence $\times$ frame}). Samples are the number of Phase-Field simulations, the sequence is the number of time observations while the frame is the RGB microstructural image. The architecture of the dataset as well as some examples proving the big variety of shapes, are shown in Fig. \ref{fig:dataset}. 
\subsection{Reduction methods}
\subsubsection{Auto-encoder neural network}
The auto-encoder is a neural network architecture for dimensionality reduction. It is trained by copying its own inputs to its labels. Starting from the original dataset composed of RGB images, the image would be compressed to a code which represents the output of the encoder. Inverse transformation could be obviously ensured. In other words, the input data are reduced into a low dimensional latent space using neural networks. Two operation should be distinguished ; the encoding process consists in mapping the input data to a latent space while the decoding step reconstructs the original image. The neural network is then splitted into the encoded and decoded segments as follows. 
\begin{equation}
\begin{aligned}
&\phi: ~\chi ~\rightarrow ~ \mathcal{F} \\
&\psi:~ \mathcal{F}~\rightarrow ~\chi  \\
&\phi, \psi ~= \argmin_{\phi,\psi}  ~ \|X - (\psi ~ \circ ~ \phi~ \circ ~X) \|^{2}
\end{aligned}
\label{eq:encoder_decoder}
\end{equation}
here $\phi$ is the encoded function which maps the input dataset \textit{X} to the latent space $\mathcal{F}$. $\psi$ is the decoded function which maps the latent space $\mathcal{F}$ to the original shape. For the encoding process, the encoder is a single-layer neural network which takes the input {\textit{x}} ($\in$ $\chi $) and maps it to the latent h ($\in$ $\mathcal{F} $) :
\begin{equation}
\begin{aligned} 
\textbf{h} =\sigma (\textbf{Wx +b} )
\end{aligned}
\label{eq:encoder}
\end{equation}
where \textit{h} denotes latent representation/variables or simply the code and $\sigma$ is the activation function such as rectified linear unit, hyperbolic tangent or sigmoid function \cite{Fetni2021esaform}. W and b are respectively the different weights and bias.
\newline Similarly, the decoder maps \textit{h} to a reconstructed input {\textit{x'}}:
\begin{equation}
\begin{aligned} 
\textbf{x'} =\sigma ' (\textbf{W'h +b'})
\end{aligned}
\label{eq:decoder}
\end{equation}
where W' and b' are respectively the different weights and bias and $\sigma$' is the activation function.
\newline The auto-encoder is then trained to minimize the reconstruction errors (known as the loss function), which is expressed as a function of the parameters in equations \ref{eq:encoder} and \ref{eq:decoder} as follows:
\begin{equation}
\begin{aligned} 
 {\mathcal {L}}(\mathbf {x} ,\mathbf {x'} )=\|\mathbf {x} -\mathbf {x'} \|^{2}=\|\mathbf {x} -\sigma '(\mathbf {W'} (\sigma (\mathbf {Wx} +\mathbf {b} ))+\mathbf {\textbf{b'}} )\|^{2}
\end{aligned}
\label{eq:loss_autoencoder}
\end{equation}
\newline In order to apply auto-encoding, different trials were conducted under High Performance Computing (HPC). First, the 4D input matrix in converted to a 2D one by flattening each input image as follows: $\{128,128,3\} \rightarrow (49152,)$ .
 Depending on the memory consumptions and computing time, the targeted size/shape for the output code range between (200,) and (2000,). This was chosen to ensure a compromise between an accurate dimensionality reduction and an easy training a time-series neural network using the obtained codes. 
Once the first reduction is performed, the input matrix ([number of samples $\times$ flattened image shape] , noted hereafter \textit{dataset}$\chi$ [\textit{number of samples} $\times$ \textit{input shape}]), is transformed to a reduced shape matrix of dimension : [\textit{number of samples} $\times$ \textit{code1}] ; here \textit{code1} is the output code size while \textit{number of samples} is the size of the dataset and is equal to [7300 $\times$ 100]. The encoded dataset is called hereafter \textit{dataset}$\mathcal{F}$1.
\par Trials on HPC clusters showed that adding hidden layers to the auto-encoder architecture requires more memory consumption as well as computing time. For that, we applied the auto-encoder differently by adding a second auto-encoder with the purpose to improve the reduction ratio given by the first one. It was expected that, through the second auto-encoder, we get a more reduced dataset denoted hereafter \textit{dataset}$\mathcal{F}$2 whose the shape is [\textit{number of samples} x \textit{code2}].
\subsubsection{Principal Component Analysis}
PCA is considered as one of the most popular algorithms for dimensionality reduction \cite{KERR2013103,SUN202294}. Indeed, as a multivariate statistical analysis method, it has the ability to convert multiple indicators, through an orthogonal linear transformation of the data to new coordinate system, into a few comprehensive indicators  \cite{SUN202294}. Here, the greatest variance lies in the first row. For a given dataset \textit{Z} of n samples of \textit{m} dimensional vectors, the covariance matrix could be expressed as:
\begin{equation}
C_{z}=\frac{1}{n-1} (Z-\overline{Z})(Z-\overline{Z})^{T}
\label{eq:pca_cov}
\end{equation}
where $\overline{Z}=\frac{1}{n}\sum_1^n Z_{i}$. $Z_{i}$ is a sample from Z.
Once data is standardized (e.g. all values range between 0 and 1) and the symmetric covariance matrix is computed, the eigenvectors and eigenvalues are extracted from it to identify for principal components of the data.
Comparing to auto-encoders, PCA can be performed faster and also computationally cheaper than training an additional layer of auto-encoders.
\newline This was confirmed by first trials on \textit{dataset}$\chi$. However, starting from the \textit{dataset}$\mathcal{F}$1 ; while the original dimension is already compressed and the targeted reduction ratio is small (1/10 or less), the hypothesis that PCA could deal with this challenge is to be investigated. In this work, PCA was applied in the second step of dimensionality reduction.
\subsection{LSTM and GRU neural network for time-series predictions}
In order to get a solution for the short-term memory and avoid the issue of vanishing gradients of RNN, LSTM architecture is based on three internal mechanisms (called gates : forgot, input and outputs gates) as well as a cell state. The flow of information can thus be regularized by considering only relevant information. GRU neural networks have similar architecture but with few differences (a reset gate instead of the forgot one and an update gate) \cite{ARUNKUMAR20227585,GAO2020125188}.
\newline The LSTM training requires a 3D matrix (\textit{samples} $\times$ \textit{time steps} $\times$ \textsl{features}) as input. \textit{Samples} corresponds to the number of Phase-Field simulations (7300 in this work). \textit{time steps} corresponds to the number of time observations : 100 in this work, while \textit{features} is the number of dimensions that are feed  to the model at each time step ; that corresponds to \textit{ (code2,)} ; a size to which we add each Phase-Field simulation characteristics (\textit{X}, \textit{M}, $\kappa$) in order to allow neural networks better correlating the data.
\section{Results and Discussions}
\subsection{Dimensionality reduction using auto-encoders and Principal Component Analysis}
The first step of reduction is ensured by a first auto-encoder (auto-encoder 1). The second step is performed using either an auto-encoder or PCA.
\subsubsection{The first step of reduction using Auto-encoders}
\begin{figure}
\centering
\subcaptionbox{ \label{}}
{\includegraphics[width=8cm,height=5cm]{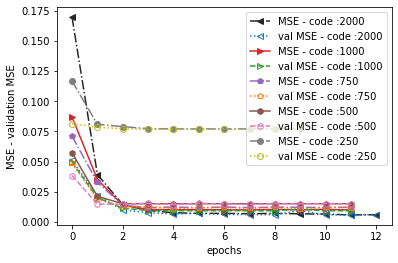}}

\subcaptionbox{ \label{}}
{\includegraphics[width=7cm,height=4cm]{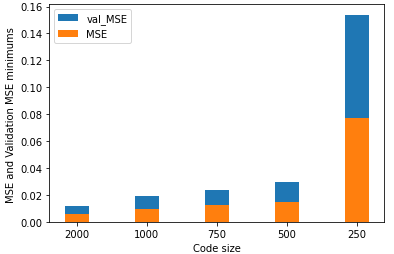}}

\caption{Results of the training of auto-encoders with different code sizes (2000, 1000, 750, 500 and 250 respectively). (a) MSE and validation MSE evolution with epochs, (b) Histogram plot for the minimum MSE and validation MSE.}
\label{fig:layer_1_autoencoder}

\end{figure}

\begin{figure}
\centering
\subcaptionbox{ \label{}}
{\includegraphics[scale=0.25]{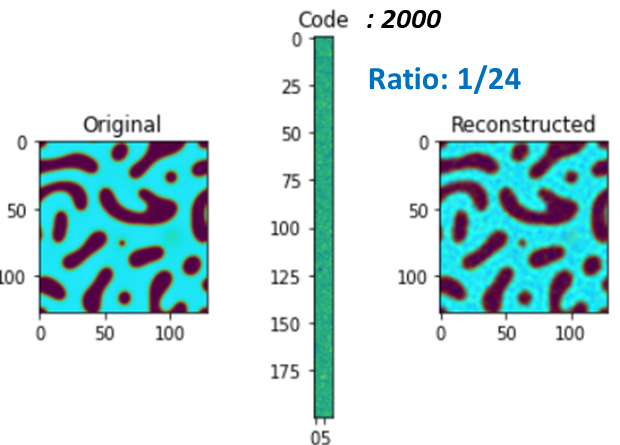}}
\subcaptionbox{ \label{}}
{\includegraphics[scale=0.75]{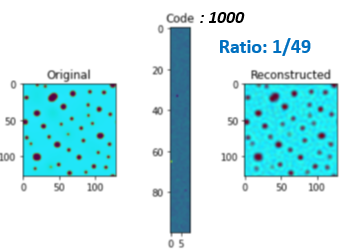}}
\subcaptionbox{ \label{}}
{\includegraphics[scale=0.25]{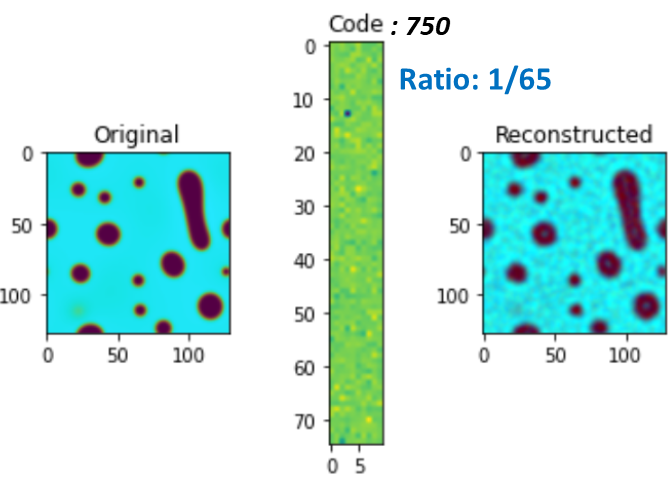}}
\subcaptionbox{ \label{}}
{\includegraphics[scale=0.45]{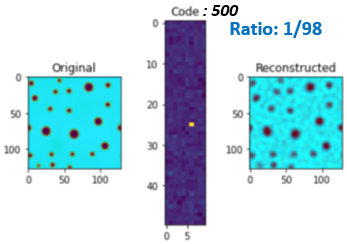}}

\caption{Reconstructed images obtained from different saved auto-encoders models. }
\label{fig:original_code_reconstructed}
\end{figure}

The results of the training of auto-encoders with different code sizes (2000, 1000, 750, 500 and 250 respectively) are shown in Fig. \ref{fig:layer_1_autoencoder}. Once trained, the models are saved in order to be used for further development. Indeed, the trained and saved models are required to encode a given RGB image and reconstruct it on one hand, and create the whole \textit{dataset}$\mathcal{F}$1 by encoding \textit{dataset}$\chi$ on the other hand. The chosen loss metric to control the convergence of the training is the Mean Square Error (MSE).
Here, the obtained results clearly show the capability of auto-encoders to get near-net shape images from the original dataset. As could be inferred from Fig. \ref{fig:layer_1_autoencoder}, the performances of reduction of the shape from ({128 $\times$ 128 $\times$ 3}) to 2000, 1000, 750 and 500 (corresponding to reduction ratio of 1/24, 1/49, 1/65 and 1/98 respectively) are similar.  
\newline Reconstructed images obtained from the different decoders are then illustrated in Fig. \ref{fig:original_code_reconstructed}. They confirm the good reduction obtained by the auto-encoders. It should be noted that a small denoising process was included in the numerical framework to improve reconstructed images.
However, for a target code with size 250, the MSE significantly increases and as a result the reconstructed obtained images are not as good compared to those obtained from higher code sizes. While a such dimension is highly needed to facilitate LSTM/GRU training, a further step of dimensional reduction is added to the framework.
\subsubsection{The second step of reduction using auto-encoders}
\par
\begin{figure}
\centering
\subcaptionbox{ \label{}}
{\includegraphics[scale=0.65]{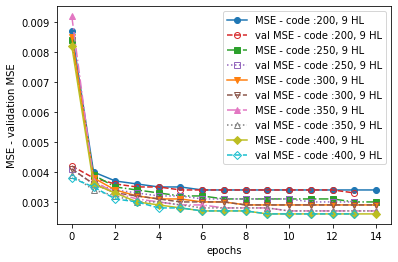}}
\subcaptionbox{ \label{}}
{\includegraphics[scale=0.5]{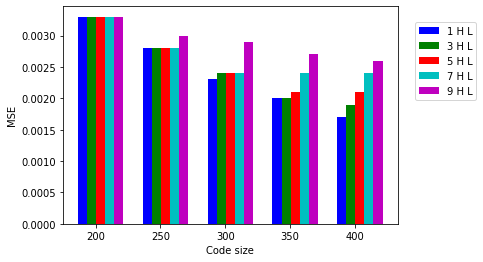}}
\caption{Training results of the 2$^{nd}$ auto-encoder for various shapes and number of hidden layers (HL) ; (a) evolution of the MSE and validation MSE for 9 HL and different code sizes. (b) MSE and validation MSE for different number of HL and code sizes.}
\label{fig:autoencoder2}
\end{figure}

 Starting from the chosen optimum code size (750,), it is interesting to investigate the effect of an additional auto-encoder to further compress images. It is reminded that this choice is due to the fact that the first encoding step requires a big amount of memory ; adding more hidden layers resulted in big need of memory in HPC computing (more than 1TB of RAM). Therefore, this is equivalent to apply an auto-encoder with 2 hidden layers.
Basing on the work of de Oca Zapiain et al. \cite{SMontes2021} and several trials in HPC, we found that a code size less than (350,) could ensure the possibility to easily train a time-series neural network when taking into account the number of samples (7300 Phase-Field simulations). In addition, when the code size is below this range, we found that the training could be performed with an ordinary computing machine (e.g. i7-10510U. 8 MB cache, up to 4.90 GHz).
Details about hardware requirement as well as computing time for each operation are given hereafter (subsection \ref{comresources}) .
In order to tune the hyper-parameters of the second auto-encoder, different optimizers were tried in a first order : sgd, adam, RMSprop and Adadelta. No significant impact of the optimizer choice on the loss evolution was found.
Sensitivity analysis of auto-encoding to the number of hidden layers and the code size (latent dimension) was conducted by progressively increasing them. The latent dimension was indeed chosen between (200,) and (400,) by a step of 50, while the number of hidden layers is increased up to 9.  
Results are then gathered in Fig. \ref{fig:autoencoder2}. It is found that, after about 10 epochs of training (Fig. \ref{fig:autoencoder2}a), the MSE reaches a plateau value of 3.10$^{-3}$. By applying different tests on the reconstructed images from the built second auto-encoder, it was observed that image quality does not reach those shown in Fig. \ref{fig:original_code_reconstructed}. The outputs images reconstructed from the first decoder of the 2$^{nd}$ auto-encoder, could be qualified from a qualitative point of view,  as better compared to the result presented in \cite{SMontes2021}. In Fig. \ref{fig:autoencoder2}b, the influence of the number of HL is highlighted ; it is obvious that increasing \textit{code2} size lead to decrease MSE, but no clear tendency could be concluded when combining code size and HL number. That may be justified by the fact that the auto-encoder architecture includes various hyper parameters in addition to the three ones investigated : number of nodes per layer, optimizer and loss function. Therefore, when dealing with a global tendency, all these hyperparameters should be analysed. The plateau reached despite the increase in the number of HL (up to 9) could reflect the limitation of using only one technique of dimensionality reduction. To address this issue, a combination of two techniques of reduction should be investigated. 
\newline Once the output image fed Phase-Field, this one has the capability of rapid regularization and smoothing out of the microstructure as it evolves. However, it is reminded that the purpose of this work is to present a framework not only limited to Phase-Field simulations, but also could be applied in other area of researches. Thus, thorough investigations were made using PCA. The purpose is to investigate the capability to ensure the second step of reduction instead of auto-encoder.
\subsubsection{The second step of reduction using Principal Component Analysis}
\begin{figure}
\centering
{\includegraphics[width=6cm,height=3.5cm]{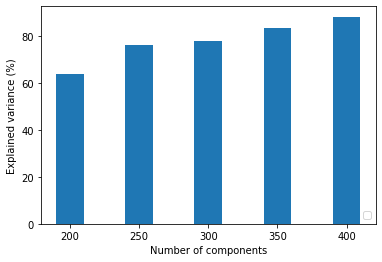}}
\caption{Explained variance obtained by the application of PCA on \textit{dataset}$\mathcal{F}$1 with different principal components.}
\label{fig:pca_hist}
\end{figure}

The hypothesis of the capabilities of PCA of giving accurate dimensionality reduction is here investigated. It is important to remind that the PCA was applied to the output shape obtained from the first layer of auto-encoders (750,). Indeed, the Phase-Field images have highly non-linear representations and PCA reduction, through its linear fitting, could not reduce a dataset of big code size (7300 x {128x128x3} ; 7300 x (49152,)). However, once the original dimension {128x128x3} is already reduced to (750,) using the first layer of reduction (auto-encoder 1), one could expect the possibility of linear representation of a relatively small code into a more reduced latent space. 
For that, PCA decompositions were conducted with different numbers of components corresponding of different reduction ratios. The histograms in Fig. \ref{fig:pca_hist} correspond to the explained variance reached for each number of principal components. Note here that before applying PCA to \textit{dataset}$\mathcal{F}$1, standardization was applied using MinMaxScaler from Scikit-Learn preprocessing library. An explained variance close to 90 \% is reached using 400 PCA components which could assimilated to a halving of the latent dimension (750,). For 250 PCA components, it is about 80 \% which could be acceptable as a compromise between the dimensionality reduction accuracy and an appropriate latent dimension for time-series analysis. 
 Similar results were obtained when trying other scalers from the same library: standardScaler, RobustScaler, PowerTransformer, QuantileTransformer and Normalizer. For MinMaxscaler, all data were ranged from 0 to 1, so transformed to the same scale. Mathematically, that is ensured for each variable \textit{X} from the dataset by : $X_{sc}=\frac{X - X_{min}}{X_{max} - X_{min} }$, where $X_{sc}$ is the scaled variable, \textit{X}$_{min}$ and X$_{max}$ are the minimum and maximum values of the dataset X. The standardization/normalization of the data consists, from a mathematical point of view, in computing the mean and the variance of each of the features present in the dataset. Then all the features are transformed as follows: $X'=\frac{X-\mu}{\sigma}$, where $\mu$ and $\sigma$ are the mean and variance of the considered population (\textit{dataset}$\mathcal{F}$1). Such a method is also called the z-score method. 
\newline

\begin{figure}
\centering
{\includegraphics[width=7cm,height=4cm]{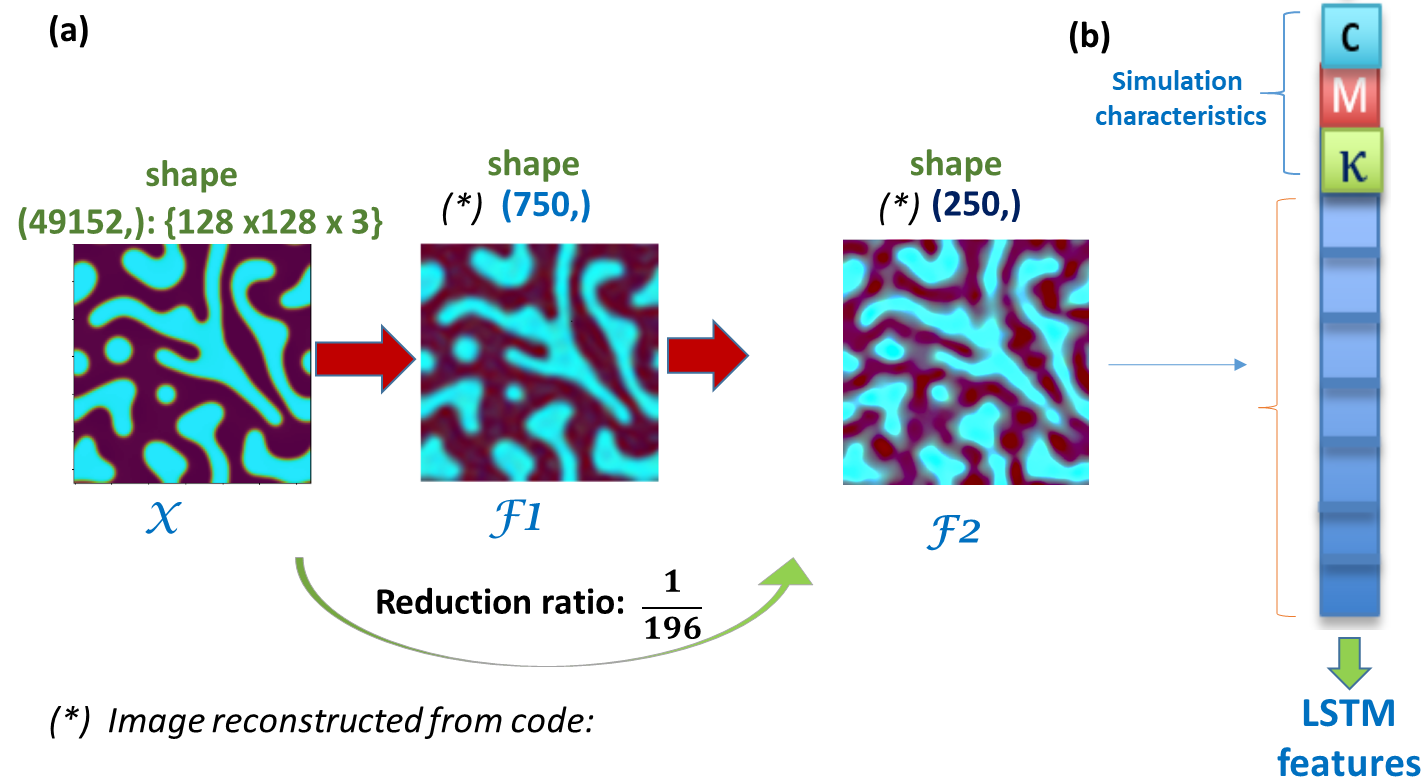}}
\caption{(a) Illustration of the flow of dimensionality reduction. Near net-shape images could be obtained from the reduced dimensions. (b): The form of the LSTM (or equivalent time series algorithm). }
\label{fig:pca_results}
\end{figure}
The combination of different methods of reductions tends to be a promising approach to handle big datasets and reach enough latent dimension to easily train recurrent neural networks. Due to the few studies in the literature focusing on Phase-Field datasets \cite{SMontes2021,HU2022115128}, it is quite difficult to establish perfectly fair comparisons between the chosen methods.
Hu et al. \cite{HU2022115128} compared the Root Mean Square Error (RMSE) obtained for PCA and UMAP: 1.64 and 8.82 respectively. But they did not put the emphasis on the accuracy after reconstruction using Phase-recovery algorithm for microstructure reconstruction from its autocorrelation representation. Similarly, \cite{SMontes2021} focused on the accuracy of LSTM predictions. 
Our results about the combination PCA auto-encoders are, for the best of our knowledge, among the first ones giving possible ratios of reductions with associated losses. 


\subsection{Time-series predictions}
\par
\begin{figure}[htbp]
\centering
\subcaptionbox{Loss and validation loss on training LSTM and GRU on the prediction of the last 5 frames basing on a based-Phase-Field images sequence. \label{}}
{\includegraphics[scale=0.65]{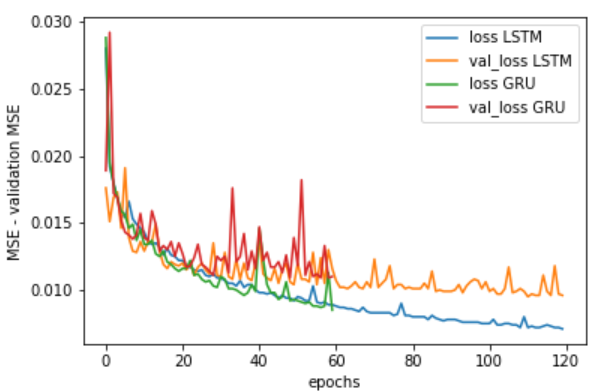}}
\subcaptionbox{Next 5 frames prediction : Phase-Field versus LSTM.\label{}}
{\includegraphics[scale=0.45]{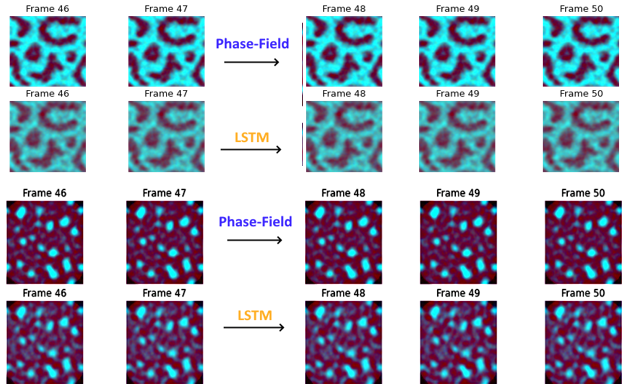}}
\subcaptionbox{Sensitivity analysis conducted to analyse the capability of LSTM to predict more frames.\label{}}
{\includegraphics[scale=0.45]{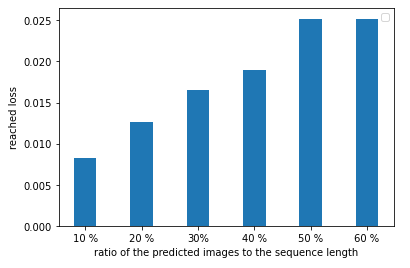}}
\caption{Results of training of time series algorithms; an emphasis was put on LSTM predictions. }
\label{fig:loss_lstm_gru}
\end{figure}

 In this part, results of times-series analysis using principally LSTM neural networks (as shown in Fig. \ref{fig:pca_results}b) are presented. The purpose is to check the capability of such time-series neural networks to make 'next-frames' predictions based on their training on the Phase-Field datasets. Training samples from the latent space $\mathcal{F}$2 are reshaped to 3D (\textit{numbers of samples $\times$ time observations $\times$ features}) as shown in Fig. \ref{fig:loss_lstm_gru}. It should be noted that the latent space $\mathcal{F}$2 is the result of a double encoding using the first auto-encoder and PCA (number of components =250) with a standard scaling in order to ensure a fast reduction. In the second stage of dimensionality reduction, PCA is chosen after a qualitative comparison of reconstructed images.
Computing were indeed conducted with the same conditions with regards to LSTM and GRU for comparison purposes. The reduced dimensionality allows to build a simplified architecture consisting in 2 layers with 500 memory blocks for each. An early stopping criterion was applied when the validation loss reaches a plateau.
\par It is reminded that the simulations characteristics (\textit{X}, \textit{M}, $\kappa$) are added to the input features to allow the model a better learning of data features. 24 hours and 56 minutes of training of the LSTM allows to reach a validation loss of 0.0082. As shown in Fig. \ref{fig:loss_lstm_gru}b, LSTM could predict the last 5 frames with high fidelity compared to Phase-Field. Once the model is trained and saved, predictions are ensured in fractions of second. It is here noted that, for simplification purposes, the sequence length was reduced from 100 to 50 by keeping the time observations with pair indices.  
It is noticed that GRU can not give the same level of loss comparing to LSTM. 
This is explained by the architecture of the two neural networks : GRU is in fact less complex by using two gates (reset and update) while LSTM uses three ones (input, forgot and output). That makes LSTM more appropriate to apply on relatively long sequences such is the case here. Meanwhile, GRU is preferred to apply on small sequences \cite{ARUNKUMAR20227585}. 
Then, a sensitivity analysis was conducted to investigate the capability of LSTM to predict more frames. This was achieved by progressively increasing the number of predicted frames. Thus, the histograms in Fig. \ref{fig:loss_lstm_gru}c correspond to the ratio of the predicted images to the sequence length (total number of images per sequence). 
Loss obviously increases with the number of outputs. But, LSTM hyperparameters could be optimized such as the number of layers, learning rate, hidden size and number of epochs.
\par It should be highlighted that obtained results were impacted by the big variety of the dataset. Indeed, varying the mobility and the gradient energy coefficient leads to completely different shapes and morphologies. For that, it is expected that fine-tuning the hyperparameters of the different neural networks of the framework (essentially auto-encoders and LSTM) could lead to improved results. These results could give insights to investigate the capacities of neural networks in resolving time-depending problems in materials science. The last frame predictions could be applied to model crack propagation. Another interesting application of times-series analysis, is the time jumping as illustrated in Fig. \ref{fig:flowchart}c ; this is not here investigated (once the emphasis was put on dimensionality reduction) but the proposed framework and the developed codes technically take into account the possibility of such analysis. The main motivation of accelerating simulations is to escape a given temporal interval when meso-scale phenomena need very fine time steps to be accurately taken into account. For that, learning the spatio-temporal behaviour of the material through the abundant experiments by literature could be a promising path to deal with the computationally expensive simulations. That requires to deal with unstructured heterogeneous data sources which is challenging but not a limitation for neural networks \cite{WU2021104066}. 
\newline We have demonstrated that auto-encoders have promising capabilities for an accurate low-dimensional non-linear representation of microstructural images. It should be highlighted that other algorithms can be investigated for dimensionality reduction such as Laplacian eigenmaps \cite{POSPELOV2021100035}, Latent map Gaussian processes \cite{OUNE2021114128} and kernel PCA \cite{BENCHEIKH2020104091}.
For time-series predictions, Hu et al. \cite{HU2022115128} recently compared RNN with LSTM and GRU in terms of accuracy and performance. They found that LSTM and GRU provide comparable accuracy with an acceptable level of fidelity. This could be explained by the few number of cells implemented in the RNNs. 
Although the frameworks proposed in this work, \cite{HU2022115128} and \cite{SMontes2021} are promising and give new insights to accelerate Phase-Field simulations, obviously Phase-Field is still needed to generate the dataset required to train neural networks, and to evolve the microstructure with two possibles cases as illustrated in Fig. \ref{fig:flowchart}c, but with a limited number of frames with reasonable accuracy (up to 10 frames). In order to achieve better Forecasting, more combinations of dimensionality reduction techniques and time series predictions should be explored. Deep transformer models for time series predictions could be explored and could be faster than RNNs and have the capability to successfully extract correlations among the elements in a long sequence \cite{WEI2022104574}. 

\subsection{Computational resources}  \label{comresources} 
Required computational resources to reproduce this work are gathered in Table \ref{tab:tab1}. As could be inferred, only the reduction of the dataset to the first latent space $\mathcal{F}$1 requires HPC resources. Once the first dataset is encoded, then further computing could be conducted under local machines. However, this is still indicative because memory would vary with the hyper parameters choice such as the number of hidden layers and the batch size.
\begin{table*}[h]
\begin{tabular}{|l|l|l|l|}
\hline
\textbf{Task}                                                                                                                                                         & \textbf{Resources}                                                                                               & \textbf{Indicative Memory usage} & \textbf{Indicative running time}                                                                                                           \\ \hline
\begin{tabular}[c]{@{}l@{}}1$^{st}$ dimensionality reduction\\ using auto-encoders\\ ($\chi$  $\rightarrow$ $\mathcal{F}$1 )\end{tabular}                      & \begin{tabular}[c]{@{}l@{}}HPC\\ single 32-core AMD \\ Epyc 7551P CPU at 2.0 GHz\\ 512 GB of RAM\end{tabular}    & up to 500 GB RAM                 & \begin{tabular}[c]{@{}l@{}}1 day, 23 hours, 15 minutes\\ ( to encode $\chi$ to the\\  latent space \\ $\mathcal{F}$1: (1000,)\end{tabular} \\ \hline
\begin{tabular}[c]{@{}l@{}}2$^{nd}$ dimensionality reduction\\  using auto-encoders\\  ($\mathcal{F}$1 $\rightarrow$  $\mathcal{F}$2)\end{tabular}                    & \begin{tabular}[c]{@{}l@{}}Ordinary computer.. e.g. \\ i7-10510U. 8 MB cache, \\ up to 4.90 GH  (*)\end{tabular} & up to 8 GB                       & \begin{tabular}[c]{@{}l@{}}few fours (depending on the \\ hyperparameters)\end{tabular}                                                    \\ \hline
\begin{tabular}[c]{@{}l@{}}2$^{nd}$ dimensionality reduction\\  using PCA\\ ($\mathcal{F}$1 $\rightarrow$  $\mathcal{F}$2)\end{tabular} & idem (*)                                                                                                          & up to 8 GB                       & \begin{tabular}[c]{@{}l@{}}few minutes depending on the \\ number of principal components\end{tabular}                                     \\ \hline
LSTM /GRU training                                                                                                                                                    & idem (*)                                                                                                          & up to 24 GB                      & \begin{tabular}[c]{@{}l@{}}24 hours, 51 minutes \\ (to train PCA for 130 epochs to \\ predict last 5 frames)\end{tabular}                  \\ \hline
\end{tabular}
\caption{Computational resources to execute the different tasks of the proposed frameworks and different sensitivity analysis}
\label{tab:tab1}
\end{table*}
\section{Conclusions}
In this work, a framework grouping auto-encoders and PCA  allowing a practical dimensionality reduction of simulated microstructural images  was presented. Such a framework gives insights to conduct time-series analysis, in particular to deal with accelerating simulations and materials failure predictions.
The most important conclusions about this work are as follows:
\begin{itemize}
	\item Auto-encoders neural networks are able to deal with microstructural-based images to ensure accurate dimensionality reduction.
	\item Auto-encoders could be associated with PCA to ensure dimensionality reduction in an efficient way.
	\item A novel framework is proposed with various gateways to transform the original dataset into a latent space. 
	\item The reduced datasets allows conducting times-series analysis using practical algorithms such as LSTM and GRU. 
\end{itemize}
For further analysis and model developments, it is suggested to do analysis on more specific datasets. e.g. by the subtraction of subsets corresponding to finer concentration ranges from the original dataset, in order to allow the neural network better learning the data features.
\section{Data availability }
Datasets related to this article as well as Python scripts can be found at \url{https://gitlab.uliege.be/S.Fetni/datasets-for-the-auto-encoder_lstm-project}, hosted at GitLab ULiège. 
\section*{Acknowledgements}
\par As Research Director of FRS-FNRS, AM Habraken acknowledges the support of this institution. 
\newline The ULiège research council of Sciences and Techniques is acknowledged for the post-doc IN IPD-STEMA 2019 grant of Seifallah Fetni.
\newline Computational resources have been provided by the Consortium des Équipements de Calcul Intensif (CÉCI), funded by the Fonds de la Recherche Scientifique de Belgique (F.R.S.-FNRS) under Grant No. 2.5020.11 and by the Walloon Region. A special thank to Mr. David Colignon for his availability and great support to successfully achieve computational tasks. 


\bibliographystyle{elsarticle-num}
\bibliography{biblio}







\end{document}